\def\paperlanguage{}
\pdfoutput=1 % for arxiv

\documentclass[letterpaper, 10 pt, conference]{ieeeconf}  % Comment this line out if you need a4paper

\usepackage{bm}
\usepackage{cite}
\usepackage{flushend}
\include{preamble}

\IEEEoverridecommandlockouts                              % This command is only needed if
\title{\LARGE \textbf
  {
    \switchlanguage%
    {%
      Motion Modification Method of Musculoskeletal Humanoids by Human Teaching Using Muscle-Based Compensation Control
    }%
    {%
      筋補償制御を用いた人間の教示による筋骨格ヒューマノイドの動作修正手法
    }%
  }
}

\author{Kento Kawaharazuka$^{1}$, Yuya Koga$^{1}$, Manabu Nishiura$^{1}$, Yusuke Omura$^{1}$, Yuki Asano$^{1}$\\ Kei Okada$^{1}$, Koji Kawasaki$^{2}$, and Masayuki Inaba$^{1}$% <-this % stops a space
  \thanks{$^{1}$ The authors are with the Department of Mechano-Informatics, Graduate School of Information Science and Technology, The University of Tokyo, 7-3-1 Hongo, Bunkyo-ku, Tokyo, 113-8656, Japan.
    {\texttt\small [kawaharazuka, koga, nishiura, omura, asano, k-okada, inaba]@jsk.t.u-tokyo.ac.jp}
  }
  \thanks{$^{2}$ The author is associated with TOYOTA MOTOR CORPORATION.
    {\texttt\small koji\_kawasaki@mail.toyota.co.jp}
  }
}
\begin{document}

\maketitle
\thispagestyle{empty}
\pagestyle{empty}

%%%%%%%%%%%%%%%%%%%%%%%%%%%%%%%%%%%%%%%%%%%%%%%%%%%%%%%%%%%%%%%%%%%%%%%%%%%%%%%%
\begin{abstract}
  \switchlanguage%
  {%
    While musculoskeletal humanoids have the advantages of various biomimetic structures, it is difficult to accurately control the body, which is challenging to model.
    Although various learning-based control methods have been developed so far, they cannot completely absorb model errors, and recognition errors are also bound to occur.
    In this paper, we describe a method to modify the movement of the musculoskeletal humanoid by applying external force during the movement, taking advantage of its flexible body.
    Considering the fact that the joint angles cannot be measured, and that the external force greatly affects the nonlinear elastic element and not the actuator, the modified motion is reproduced by the proposed muscle-based compensation control.
    This method is applied to a musculoskeletal humanoid, Musashi, and its effectiveness is confirmed.
  }%
  {%
    筋骨格ヒューマノイドには様々な生物模倣型の利点がある一方, そのモデル化困難な身体を正確に制御することは難しい.
    これまで様々な学習型制御手法が開発されてきたが, その誤差を完全に吸収することはできず, 認識等の誤差もつきまとう.
    そこで本研究では, 柔軟な身体の特徴を活かし, ロボット動作中に人間が外力を加えることでその動作を意図したように修正する手法について述べる.
    通常の軸駆動型ロボットと違い関節角度が測定できないことや, 外力がアクチュエータでなく非線形弾性要素側に大きく影響する点などを踏まえた, 筋補償制御による修正された動作の再生を提案する.
    本手法を筋骨格ヒューマノイドMusashiに適用し, その有効性を確認する.
  }%
\end{abstract}

\section{INTRODUCTION}\label{sec:introduction}
\begin{figure}[t]
  \centering
  \includegraphics[width=1.0\columnwidth]{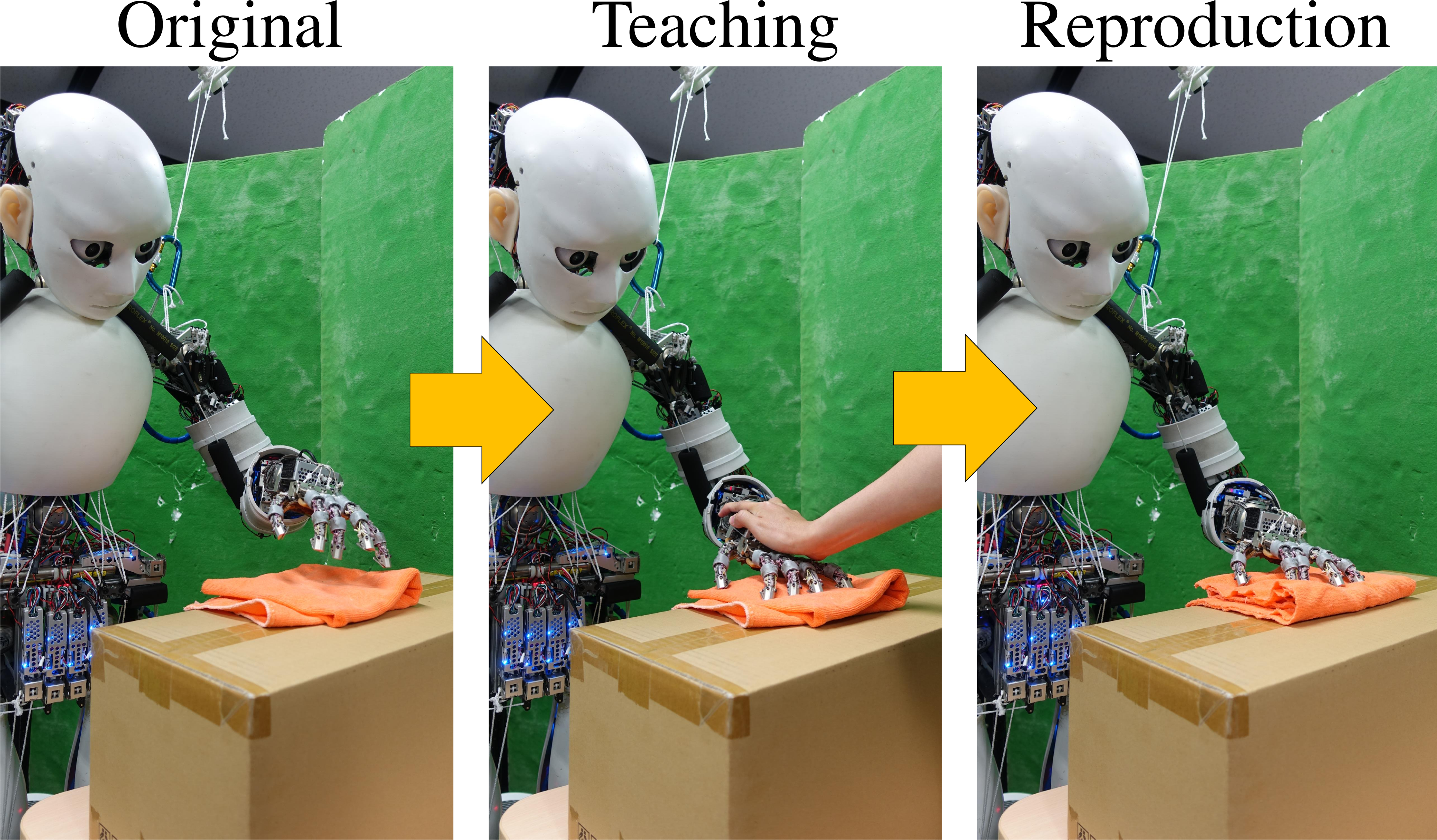}
  \vspace{-3.0ex}
  \caption{The overall flow of this study: teaching by humans during the original movement and its reproduction.}
  \label{figure:motivation}
  \vspace{-3.0ex}
\end{figure}

\switchlanguage%
{%
  The musculoskeletal humanoid \cite{nakanishi2013design, wittmeier2013toward, jantsch2013anthrob, asano2016kengoro} has biomimetic advantages such as the redundant muscle arrangement, variable stiffness control using nonlinear elasticity and antagonism, ball joints without singular points, and the flexible spine and fingers.
  At the same time, the complex and flexible body is difficult to model, and various learning control methods have been developed in order to achieve accurate movements.
  In \cite{mizuuchi2006acquisition}, joint angles obtained from motion capture and muscle lengths are correlated offline using a neural network.
  In \cite{ookubo2015learning}, joint angles obtained from IMU and muscle lengths are correlated using polynomial regression.
  In \cite{kawaharazuka2018online, kawaharazuka2018bodyimage, kawaharazuka2019longtime, kawaharazuka2020autoencoder}, the relationship among joint angles obtained from vision, muscle tensions, and muscle lengths is learned online using a neural network.
  While these control methods are able to realize the intended joint angles to some extent, because they handle only static models, there are always some errors due to hysteresis, inter-skeletal friction, etc.
  In response to this problem, some methods have been developed to handle dynamic models, such as \cite{kawaharazuka2019pedal}, but they cannot handle time series information in a large space of joint angles, muscle tensions, and muscle lengths.
  In addition, recognition errors in the robot are common during motion generation, and even if the robot can move accurately, it may not be able to perform the task accurately.

  Therefore, in this study, we change the approach and consider to modify the original motion by applying external force during the motion and then reproduce the modified motion.
  Humans can easily change the movements of musculoskeletal humanoids from the outside without any controls due to their flexible bodies, and thus they are considered to have a high affinity with human teaching.
  At the same time, it is not possible to use the usual teaching methods, such as \cite{billard2008survey}, because the structure of the musculoskeletal humanoid is different from that of the conventional axis-driven humanoid.
  This is because musculoskeletal humanoids usually do not have sensors to directly measure joint angles due to the presence of ball joints and the complex scapula.
  Also, nonlinear elastic elements, which ensure the flexibility and variable stiffness control, are provided at the end of the muscle, and the displacement of motion caused by external force is transmitted to the nonlinear elastic elements rather than to the actuator side.
  Therefore, it is difficult to reproduce the modified motion in the same way with ordinary humanoids.
  Also, direct teaching methods such as \cite{mizuuchi2004kenta} and wearable devices such as \cite{kurotobi2012device} have been developed for musculoskeletal humanoids, but these methods do not take into account the fact that the robot is subjected to external force during the motion.
  In this study, we propose a method to accurately reproduce the modified motion by using muscle tensions during human teaching without using joint angle information and by performing compensatory control at the muscle level (muscle-based compensation control).
  We apply this method to the musculoskeletal humanoid, Musashi \cite{kawaharazuka2019musashi}, and confirm the effectiveness of this study by performing box wiping and drawing behaviors as well as a basic comparison experiment.
}%
{%
  筋骨格ヒューマノイド\cite{nakanishi2013design, wittmeier2013toward, jantsch2013anthrob, asano2016kengoro}は, 冗長な筋配置, 非線形弾性と拮抗を用いた可変剛性制御, 特定点のない球関節, 柔軟な背骨や指等の多数の生物模倣型の利点を有する.
  同時に, その複雑で柔軟な身体はモデル化困難であり, 正確な動作を求めてこれまで様々な学習型制御手法が開発されてきた.
  \cite{mizuuchi2006acquisition}では, motion captureから得られる関節角度と筋長センサの値がニューラルネットワークを用いて対応付けられた.
  \cite{ookubo2015learning}では, 多項式近似を用いてIMUから得られた関節角度と筋長を対応付けている.
  \cite{kawaharazuka2018online, kawaharazuka2018bodyimage, kawaharazuka2019longtime, kawaharazuka2020autoencoder}ではニューラルネットワークを用いて視覚から得られた関節角度, 筋張力, 筋長の関係をオンラインで学習している.
  これらの手法はある程度正しい指令関節角度を実現できる一方, 静的なモデルのみを扱っているため, ヒステリシスや骨格間摩擦等から, 誤差は必ず残ってしまう.
  この問題に対して\cite{kawaharazuka2019pedal}のように動的なモデルを扱う手法も開発されてはいるが, 関節角度・筋張力・筋長という大きな空間の時系列は扱うことができていない.
  また, 動作生成の際にロボットにおける認識誤差はつきものであり, 正確に動作できたとしても正確にタスクをこなせるとは限らない.

  そのため本研究ではアプローチを変え, \figref{figure:motivation}のように, 筋骨格ヒューマノイドの動作中に外力を加えることでその動作を意図したものに修正し再生する方法について考える.
  筋骨格ヒューマノイドはその柔軟な身体から, 制御なしに簡単に外から動きを変化させることができるため, 人間による教示との親和性が高いと考えられる.
  しかし同時に, 通常の軸駆動型ロボットと違う構造であるため, 通常の教示\cite{billard2008survey}手法を用いることはできない.
  これは, 通常筋骨格ヒューマノイドは球関節や複雑な肩甲骨等の存在から関節角度を直接測定するセンサがないこと, そして, 柔軟性と可変剛性制御を担保する非線形弾性が筋末端に備わっているため, 外力による動作の変位がアクチュエータ側ではなく, 非線形弾性要素側に伝達されること, の二点が大きな要因である.
  そのため, 外力を加えて修正した動作を, 同じように再生することが難しい.
  また, 筋骨格ヒューマノイドにおいても\cite{mizuuchi2004kenta}のようなdirect teachingや, \cite{kurotobi2012device}のウェアラブルデバイスを使った教示方法が開発されているが, どちらも動作中に外部から力を受けることを考慮した手法ではない.
  本研究では, 関節角度を介さず教示中の筋張力を用いて, 筋レベルでの補償制御を行うことで, 修正された動作を正確に再現する手法について提案する.
  筋骨格ヒューマノイドMusashi\cite{kawaharazuka2019musashi}に本手法を適用し, 基本的な比較実験の他ボックス拭きや絵描きの動作を行うことで, 本研究の有効性を確認する.
}%

\switchlanguage%
{%
  This study is organized as follows.
  In \secref{sec:musculoskeletal-humanoids}, the basic musculoskeletal structure and Musculoskeletal AutoEncoder \cite{kawaharazuka2020autoencoder}, a learning control method developed previously and is used for muscle-based compensation control, are described.
  In \secref{sec:proposed}, the overall flow of this study and muscle tension limiter are described, and then the proposed method of muscle-based compensation control is described.
  In \secref{sec:experiment}, a comparison of the methods described in \secref{sec:proposed} is presented, and experiments on practical tasks using the proposed method are described.
  Finally, a discussion is given in \secref{sec:discussion} and conclusions are presented in \secref{sec:conclusion}.
}%
{%
  本研究の構成は以下のようになっている.
  \secref{sec:musculoskeletal-humanoids}では筋骨格ヒューマノイドの基本構成と, 筋補償制御と併用するこれまでに開発された学習型制御手法であるMusculoskeletal AutoEncoderについて説明する.
  \secref{sec:proposed}では, 全体の流れ・安全機構について説明した後, 提案手法である筋補償制御について述べる.
  \secref{sec:experiment}では\secref{sec:proposed}で述べた手法の比較, これを用いた実用的タスクの実験について述べる.
  最後に, \secref{sec:discussion}にて議論, \secref{sec:conclusion}にて結論を述べる.
}%

\begin{figure}[t]
  \centering
  \includegraphics[width=0.9\columnwidth]{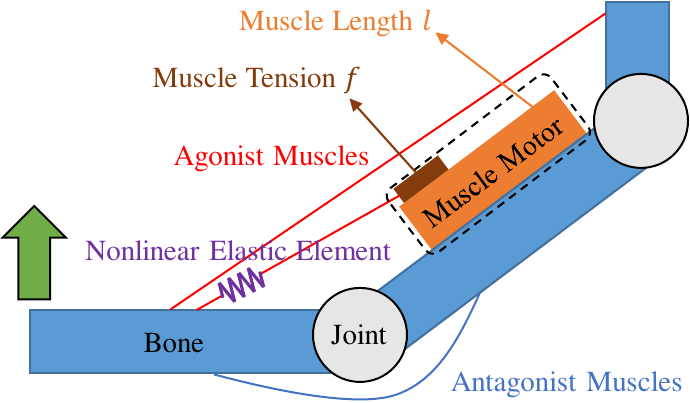}
  \caption{The basic musculoskeletal structure.}
  \label{figure:musculoskeletal-structure}
  \vspace{-3.0ex}
\end{figure}

\section{Musculoskeletal Humanoids and Musculoskeletal AutoEncoder} \label{sec:musculoskeletal-humanoids}
\subsection{The Basic Structure of Musculoskeletal Humanoids} \label{subsec:basic-structure}
\switchlanguage%
{%
  The basic structure of the musculoskeletal humanoid is shown in \figref{figure:musculoskeletal-structure}.
  In this study, we mainly handle the musculoskeletal structure in which the muscles are wound by a motor via a pulley.
  The redundant muscles are arranged antagonistically around joints.
  The muscles are mainly composed of Dyneema, which is a friction-resistant synthetic fiber, and often have a nonlinear elastic element at the end that enables variable stiffness control.
  The nonlinear elastic element is elongated by applying muscle tension, and at the same time, Dyneema itself has elasticity, creating the flexibility of the body.
  Muscle length $\bm{l}$ can be measured from an encoder attached to the motor, and muscle tension $\bm{f}$ can be measured from a muscle tension measurement unit.
  Although joint angles $\bm{\theta}$ cannot usually be measured as described in \secref{sec:introduction}, it is possible to measure them in some robots by using a special mechanism \cite{urata2006sensor, kawaharazuka2019musashi}.
  Even when joint angles cannot be directly measured, it is possible to estimate the joint angles of the actual robot using visual sensors, markers attached at the hand, and changes in muscle lengths, as in \cite{kawaharazuka2018online}, and this data can be used for learning.
  However, because of the disadvantage of having to constantly look at the end-effectors, some methods \cite{kawaharazuka2018bodyimage, kawaharazuka2019longtime, kawaharazuka2020autoencoder} have been developed to estimate $\bm{\theta}$ from $(\bm{f}, \bm{l})$ by learning the relationship among $(\bm{\theta}, \bm{f}, \bm{l})$.
}%
{%
  筋骨格ヒューマノイドの基本的な構造を\figref{figure:musculoskeletal-structure}に示す.
  本研究では主に, 筋をプーリを介してモータで巻き取る方式の筋骨格構造について扱う.
  冗長な筋が関節の周りに拮抗に配置されている.
  筋は主に摩擦に強い化学繊維であるDyneemaで構成されており, 末端に可変剛性制御を可能とする非線形弾性要素が配置されている場合が多い.
  筋張力をかけることで非線形弾性が伸びると同時に, Dyneema自体も弾性を有しており, 身体の柔軟性を生み出している.
  センサとして, モータについたエンコーダから筋長$\bm{l}$が, 筋張力測定ユニットから筋張力$\bm{f}$が測定可能である.
  \secref{sec:introduction}で述べたように通常は関節角度$\bm{\theta}$を測定できないが, 一部のロボットでは特殊な機構を用いてそれらを測定することが可能である\cite{urata2006sensor, kawaharazuka2019musashi}.
  関節角度が直接測れない場合も, \cite{kawaharazuka2018online}のように視覚センサと筋長変化を用いて実機関節角度を推定することが可能であり, 学習の際はこれを実機センサデータとして用いる.
  しかし, 常にエンドエフェクタを見続けなければならないという難点があるため, $(\bm{\theta}, \bm{f}, \bm{l})$の関係を学習することで, $(\bm{f}, \bm{l})$から$\bm{\theta}$を推定できるようにする手法が開発されている\cite{kawaharazuka2018bodyimage, kawaharazuka2019longtime, kawaharazuka2020autoencoder}.
}%

\subsection{Musculoskeletal AutoEncoder} \label{subsec:musculoskeletal-autoencoder}
\switchlanguage%
{%
  We briefly describe Musculoskeletal AutoEncoder (MAE) \cite{kawaharazuka2020autoencoder}, which is used for the muscle-based compensation control proposed in this study.
  Note that in this study, the muscle tension $\bm{T}$ and the function $\bm{f}$ in \cite{kawaharazuka2020autoencoder} are converted to $\bm{f}$ and $\bm{h}$, respectively.

  MAE is a neural network representing the relationship among $(\bm{\theta}, \bm{f}, \bm{l})$: $(\bm{\theta}, \bm{f})\to\bm{l}$, $(\bm{f}, \bm{l})\to\bm{\theta}$, and $(\bm{l}, \bm{\theta})\to\bm{f}$.
  An AutoEncoder-type network with $(\bm{\theta}, \bm{f}, \bm{l})$ and mask value as input, and $(\bm{\theta}, \bm{f}, \bm{l})$ as output is updated from the actual robot sensor information.
  Here, the functions $\bm{h}_{\bm{l}}$ and $\bm{h}_{\bm{\theta}}$, which can be derived from MAE, are defined as $\bm{l} = \bm{h}_{\bm{l}}(\bm{\theta}, \bm{f})$, $\bm{\theta}=\bm{h}_{\bm{\theta}}(\bm{l}, \bm{f})$.
  MAE is trained offline and online using the information of actual sensor data.
  The relationship among $(\bm{\theta}, \bm{f}, \bm{l})$ including the information of muscle Jacobian, nonlinear elastic behaviors, etc. is embedded into MAE.
  By using $\bm{h}_{\bm{\theta}}$ in MAE, the current estimated joint angle $\bm{\theta}^{est}$ can be calculated from the information of $(\bm{f}, \bm{l})$ (this operation is referred to as EST).
  Also, by using $\bm{h}_{\bm{l}}$, the target muscle length $\bm{l}^{ref}$ realizing the target joint angle $\bm{\theta}^{ref}$ can be calculated (this operation is referred to as CTRL).
  It should be noted that it is necessary to calculate the target muscle tension $\bm{f}^{ref}$, which is determined by an iterative calculation using the gravity compensation torque $\bm{\tau}$ and the backpropagation method \cite{kawaharazuka2020autoencoder}.
  Also, since the target value $\bm{l}^{ref}$ and the measured value $\bm{l}$ are different, the muscle stiffness control \cite{shirai2011stiffness} must be taken into account.
  Therefore, let $\bm{l}^{ref}$ be the value obtained by adding $\bm{l}^{comp}$ to $\bm{h}_{\bm{l}}(\bm{\theta}^{ref}, \bm{f}^{ref})$, as below,
  \begin{align}
    &\bm{l}^{comp}(\bm{f}) = -(\bm{f} - \bm{f}^{bias})/K \label{eq:msc}\\
    &\bm{l}^{ref} = \bm{h}_{\bm{l}}(\bm{\theta}^{ref}, \bm{f}^{ref}) + \bm{l}^{comp}(\bm{f}^{ref})
  \end{align}
  where $\bm{f}^{bias}$ is the bias term of the muscle stiffness control and $K$ is the stiffness coefficient.

  Note that since MAE represents only static intersensory relationships, it is not possible to estimate completely accurate $\bm{\theta}^{est}$ by EST due to hysteresis and friction, and it is not possible to achieve completely accurate $\bm{\theta}^{ref}$ by CTRL.
}%
{%
  本研究で提案する筋補償制御が併用するMusculoskeletal AutoEncoder (MAE) \cite{kawaharazuka2020autoencoder}について簡単に説明をする.
  なお本研究では, \cite{kawaharazuka2020autoencoder}における筋張力$\bm{T}$は$\bm{f}$に, 関数$\bm{f}$は$\bm{h}$に変換されていることに注意されたい.

  MAEは, $(\bm{\theta}, \bm{f}, \bm{l})$の間の関係である, $(\bm{\theta}, \bm{f})\to\bm{l}$, $(\bm{f}, \bm{l})\to\bm{\theta}$, $(\bm{l}, \bm{\theta})\to\bm{f}$という3つの関係を一つのネットワークで表したものである.
  入力を$(\bm{\theta}, \bm{f}, \bm{l})$とマスク値, 出力を$(\bm{\theta}, \bm{f}, \bm{l})$としたAutoEncoder型のネットワークを実機センサデータから更新する.
  ここで, MAEのから取り出せる関数$\bm{h}_{\bm{l}}$と$\bm{h}_{\bm{\theta}}$を, $\bm{l} = \bm{h}_{\bm{l}}(\bm{\theta}, \bm{f})$, $\bm{\theta}=\bm{h}_{\bm{\theta}}(\bm{l}, \bm{f})$のように定義する.
  MAEにおける$\bm{h}_{\bm{\theta}}$を用いることで, 現在の推定関節角度$\bm{\theta}^{est}$を$(\bm{f}, \bm{l})$の情報から計算することができる(以降, この操作をESTと呼ぶ).
  また, $\bm{h}_{\bm{l}}$を用いることで, ある指令値$\bm{\theta}^{ref}$を実現するための筋長$\bm{l}^{ref}$を計算することができる(以降, この操作をCTRLと呼ぶ).
  なお, ここでは筋張力の指令値$\bm{f}^{ref}$を計算する必要があるが, これは重力補償トルク$\bm{\tau}$と誤差逆伝播法を用いた反復計算により決定される\cite{kawaharazuka2020autoencoder}.
  また, 指令値$\bm{l}^{ref}$と筋長$\bm{l}$は異なるため, 筋剛性制御\cite{shirai2011stiffness}も考慮しなければならない.
  よって, 以下のように$\bm{h}_{\bm{l}}(\bm{\theta}^{ref}, \bm{f}^{ref})$に$\bm{l}^{comp}$を足しこんだ値を$\bm{l}^{ref}$とする.
  \begin{align}
    &\bm{l}^{comp}(\bm{f}) = -(\bm{f} - \bm{f}^{bias})/K \label{eq:msc}\\
    &\bm{l}^{ref} = \bm{h}_{\bm{l}}(\bm{\theta}^{ref}, \bm{f}^{ref}) + \bm{l}^{comp}(\bm{f}^{ref})
  \end{align}
  ここで, $\bm{f}^{bias}$は筋剛性制御のバイアス項, $K$は筋剛性制御の剛性係数である.

  MAEは静的なセンサ関係のみ表しているため, ヒステリシス等の問題から, ESTにより完全に正確な$\bm{\theta}^{est}$が推定できるわけでなく, CTRLにより, 完全に正確な$\bm{\theta}^{ref}$が実現できるわけでもないことに注意されたい.
}%

\section{Motion Modification Using Muscle-based Compensation Control} \label{sec:proposed}
\switchlanguage%
{%
  First of all, the overall flow of motion modification by human teaching is described below.
  \begin{enumerate}
    \renewcommand{\labelenumi}{(\alph{enumi})}
    \item Making the robot motion (original)
    \item Modifying the motion by applying external force while the motion is running (teaching)
    \item Reproducing the modified motion (reproduction)
  \end{enumerate}
  (a) is usually programmed by humans or generated from the results of recognition and so on.
  The target joint angle $\bm{\theta}^{ref}_{t}$ is determined, and CTRL calculates the target muscle length $\bm{l}^{ref}_{t}$ and target muscle tension $\bm{f}^{ref}_{t}$.
  Here, $\bm{\bullet}_t$ refers to the value at the time step $t$ ($0 \leq t < T$), where $T$ denotes the length of the motion.
  In (b), the muscle length $\bm{l}^{data}_{t}$ and muscle tension $\bm{f}^{data}_{t}$ measured during the teaching are accumulated.
  During the teaching, the original flexibility of the musculoskeletal structure can be used, but it is also possible to increase the effect of the external force by limiting the maximum muscle tension (this will be explained in \secref{subsec:tension-limiter}).
  In (c), the modified motion is reproduced by calculating the muscle length $\Delta\bm{l}^{ref}_{t}$ to be changed from $\bm{l}^{ref}_{t}$ using the obtained $\{\bm{l}, \bm{f}\}^{\{ref, data\}}_{t}$ and sending $\bm{l}^{ref}_{t}+\Delta\bm{l}^{ref}_{t}$ to the actual robot (this will be explained in \secref{subsec:compensation-control}).
  The methods of reproducing the modified motion for comparison experiments is summarized in \secref{subsec:comparison-methods}.
}%
{%
  まず, 本研究における人間の教示による動作修正の一連の流れを以下に述べる.
  \begin{enumerate}
    \renewcommand{\labelenumi}{(\alph{enumi})}
    \item ロボットの動作を作成する.
    \item 動作中に人間が外力を加えることでその動作を修正する.
    \item 修正された動作を再生する.
  \end{enumerate}
  (a)は通常人間が動きをプログラムしたり, 認識等の結果から動作を生成したりする.
  指令関節角度$\bm{\theta}^{ref}_{t}$が決定され, CTRLにより指令筋長$\bm{l}^{ref}_{t}$, 指令筋張力$\bm{f}^{ref}_{t}$が計算される.
  ここで, $\bm{\bullet}_t$はある時刻ステップ$t$ ($0 \leq t < T$)における値を指す($T$はその動作の長さを表す).
  (b)において, 動作中に測定された筋長$\bm{l}^{data}_{t}$, 筋張力$\bm{f}^{data}_{t}$を蓄積しておく.
  また, (b)では筋骨格構造の元々の柔軟性を利用することも可能であるが, 筋張力の最大値を制限することでより外力の効果を大きくすることも可能である(これは\secref{subsec:tension-limiter}において説明する).
  (c)において, 得られた$\{\bm{l}, \bm{f}\}^{\{ref, data\}}_{t}$を用いて$\bm{l}^{ref}_{t}$から変化させるべき筋長$\Delta\bm{l}^{ref}_{t}$を計算し, $\bm{l}^{ref}_{t}+\Delta\bm{l}^{ref}_{t}$を送ることで修正された動作を再生する(これは\secref{subsec:compensation-control}において説明する).
  比較実験のための動作生成方法については, \secref{subsec:comparison-methods}においてまとめる.
}%

\subsection{Muscle Tension Limiter} \label{subsec:tension-limiter}
\switchlanguage%
{%
  For each muscle, the muscle tension limiter calculates the degree of muscle length relaxation $\Delta{l}^{ref}_{e, t}$ according to the current muscle tension as below, and sends $l^{ref}_{t}+\Delta{l}^{ref}_{e, t}$ to the actual robot.
  \begin{align}
    &if\;\;f_{t} > f^{max} \nonumber\\
    &\;\;\;\;\;\;\;\;\;\;\;\;\Delta{l}^{ref}_{e, t} = \Delta{l}^{ref}_{e, t-1} + min(C_{gain}d-\Delta{l}^{ref}_{e, t-1}, C_{plus}d)\nonumber\\
    &else \nonumber\\
    &\;\;\;\;\;\;\;\;\;\;\;\;\Delta{l}^{ref}_{e, t} = \Delta{l}^{ref}_{e, t-1} + max(0-\Delta{l}^{ref}_{e, t-1}, -C_{minus}d)\nonumber\\
    &d_{t} = |f_{t}-f^{max}| \label{eq:limiter}
  \end{align}
  where $f^{max}$ is the threshold of the muscle tension $f$ that begins to relax the muscle length, $|\bm{\bullet}|$ is the absolute value, $C_{\{minus, plus\}}$ is the coefficient that determines the amount of muscle length change in one step in the negative or positive direction, and $C_{gain}$ is the coefficient that determines the maximum amount of relaxation.
  In other words, the muscle is relaxed and tensed so that the muscle tension does not exceed the maximum value while limiting $\Delta{l}^{ref}_{e, t}$ by $C_{minus}d_{t}$ and $C_{plus}d_{t}$.
  When external force is applied, the muscle tension increases, and the muscle is stretched by the amount over $f^{thre}$, so that the motion can be modified more easily and significantly by teaching.
  In this study, we set $C_{minus}=0.001$ [mm/N], $C_{plus}=0.003$ [mm/N], and $C_{gain}=2.0$ [mm/N], and this control is performed with a period of $8$ msec.
  $f^{thre}$ is varied according to the experiment.
}%
{%
  それぞれの筋について, 筋張力制限制御は以下のように筋張力に応じて筋長弛緩度$\Delta{l}^{ref}_{e, t}$を計算し, 実機には$l^{ref}_{t}+\Delta{l}^{ref}_{e, t}$を送る.
  \begin{align}
    &if\;\;f_{t} > f^{max} \nonumber\\
    &\;\;\;\;\;\;\;\;\;\;\;\;\Delta{l}^{ref}_{e, t} = \Delta{l}^{ref}_{e, t-1} + min(C_{gain}d-\Delta{l}^{ref}_{e, t-1}, C_{plus}d)\nonumber\\
    &else \nonumber\\
    &\;\;\;\;\;\;\;\;\;\;\;\;\Delta{l}^{ref}_{e, t} = \Delta{l}^{ref}_{e, t-1} + max(0-\Delta{l}^{ref}_{e, t-1}, -C_{minus}d)\nonumber\\
    &d_{t} = |f_{t}-f^{max}| \label{eq:limiter}
  \end{align}
  ここで, $f^{max}$は筋長を弛緩させ始める筋張力$f$の閾値, $|\bm{\bullet}|$は絶対値, $C_{\{minus, plus\}}$はマイナス方向またはプラス方向に対する一ステップの筋長変化量を決める係数, $C_{gain}$は最大弛緩量を決める係数である.
  つまり, $C_{minus}d_{t}$, $C_{plus}d_{t}$で制限をかけながら, 筋張力が最大値を越えないように筋を弛緩・緊張させている.
  外から力を加えた場合筋張力が高まり, $f^{thre}$を超えた分だけ筋が伸びることで, より簡単に教示により大きく動作を修正することができるようになる.
  本研究では, $C_{minus}=0.001$ [mm/N], $C_{plus}=0.003$ [mm/N], $C_{gain}=2.0$ [mm/N]とし, 本制御は$8$ msec周期で行う.
  また, $f^{thre}$は実験に応じて変化させる.
}%

\subsection{Muscle-based Compensation Control} \label{subsec:compensation-control}
\begin{figure}[t]
  \centering
  \includegraphics[width=0.8\columnwidth]{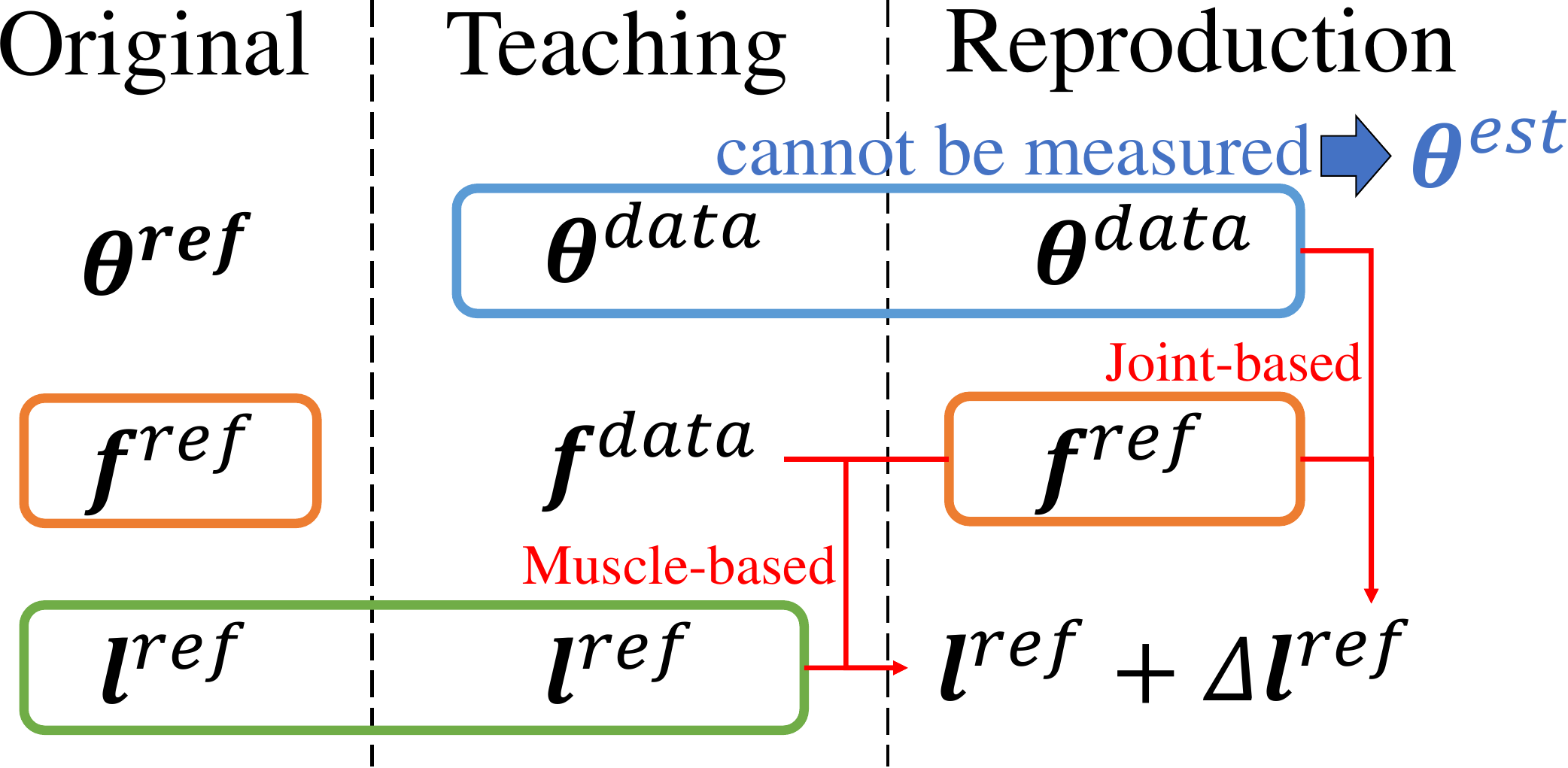}
  \caption{The relationship of $(\bm{\theta}, \bm{f}, \bm{l})$ among original, teaching, and reproduction.}
  \label{figure:relationship}
  %\vspace{-1.0zh}
\end{figure}

\switchlanguage%
{%
  For the ordinary axis-driven humanoid, we only need to take the joint angle $\bm{\theta}^{data}_{t}$ of the modified motion and send it to the actual robot as the target value $\bm{\theta}^{ref}_{t}$.
  However, such a method is impractical because musculoskeletal humanoids cannot measure the joint angle $\bm{\theta}^{data}_{t}$ directly, as described in \secref{sec:introduction}.
  The joint angle $\bm{\theta}^{est}_{t}$ estimated by EST can be used, but the error of MAE accumulates due to the two steps of estimation by EST and calculation by CTRL (the comparison experiment will be performed in the experimental section).
  On the other hand, we can take the muscle length $\bm{l}^{data}_{t}$ of the modified behavior and send it to the actual robot as the target value $\bm{l}^{ref}_{t}$.
  However, as described in \secref{subsec:basic-structure}, the effect of external force appears not on $\bm{l}^{data}_{t}$ but on the elongation of the hardware of the nonlinear elastic element or the muscle wire itself.
  Therefore, even if the measured muscle length is sent, it is not possible to reflect the external force well.
  In this study, we propose to calculate a term $\Delta\bm{l}^{ref}_{t}$ that compensates muscle length at the muscle level based on the information obtained from MAE described in \secref{subsec:musculoskeletal-autoencoder}, and reproduce the modified motion by adding it to $\bm{l}^{ref}_{t}$.

  We propose a method to reproduce the modified motion by adding the following terms (A)-(C) to the original target muscle length $\bm{l}^{ref}_{t}$.
  These are (A) $\Delta\bm{l}^{ref}_{e, t}$, the elongation due to the muscle tension limiter in \secref{subsec:tension-limiter}, (B) $\Delta\bm{l}^{ref}_{h, t}$, the elongation due to the hardware elasticity of the nonlinear elastic element or the muscle wire itself, and (C) $\Delta\bm{l}^{ref}_{s, t}$, the elongation due to the software of the muscle stiffness control.

  (A) is very simple: add $\Delta\bm{l}^{ref}_{e, t}$ in \equref{eq:limiter} to $\bm{l}^{ref}_{t}$.

  (B) calculates the muscle elongation term due to the hardware using MAE as below.
  \begin{align}
    \bm{\theta}^{est}_{t} &= \bm{h}_{\bm{\theta}}(\bm{f}^{data}_{t}, \bm{l}^{data}_{t})\nonumber\\
    \Delta\bm{l}^{ref}_{h, t} &= -(\bm{h}_{\bm{l}}(\bm{\theta}^{est}_{t}, \bm{f}^{data}_{t})-\bm{h}_{\bm{l}}(\bm{\theta}^{est}_{t}, \bm{f}^{ref}_{t})) \label{eq:hardware-compensation}
  \end{align}
  $\Delta\bm{l}^{ref}_{h, t}$ indicates the extent to which muscle length as hardware changes when changing the muscle tension at the same joint angle.
  This calculation takes advantage of the fact that (i) the muscle tension required for gravity compensation and (ii) the hardware elasticity of the muscle do not change significantly between $\bm{\theta}^{ref}$ and $\bm{\theta}^{data}$.
  Although the muscle tension is increased by the external force at the time of teaching, by assuming (i), we make sure that the muscle tension, when the modified motion is reproduced, is close to that of the original $\bm{f}^{ref}$.
  Also, the assumption in (ii) simplifies the equation of \equref{eq:hardware-compensation}.
  Indeed, $\bm{h}_{\bm{l}}(\bm{\theta}, \bm{f})$ can be decomposed into $\bm{h}_{1}(\bm{\theta})$ and $\bm{h}_{2}(\bm{\theta}, \bm{f})$, as in \cite{kawaharazuka2018bodyimage}.
  $\bm{h}_{1}$ represents the muscle length at $\bm{\theta}$ when $\bm{f}=\bm{0}$, and $\bm{h}_{2}$ represents the compensation term for the hardware elongation of the muscle length to keep $\bm{\theta}$ (always negative).
  Therefore, by substituting the same $\bm{\theta}^{est}_{t}$ for $\bm{\theta}$, $\bm{h}_{1}$ is canceled out in \equref{eq:hardware-compensation}, and the change in the compensation term for the hardware elongation of the muscle around $\bm{\theta}^{est}_{t}$, $\bm{h}_{2}(\bm{\theta}^{est}_{t}, \bm{f}^{data}_{t})-\bm{h}_{2}(\bm{\theta}^{est}_{t}, \bm{f}^{ref}_{t})$, which is only the change in the hardware elongation of the muscle, can be calculated.
  Since this is a compensation term and a negative value, it is necessary to reverse the sign of the term.

  (C) calculates the software muscle elongation from the equation of the muscle stiffness control in \equref{eq:msc} as below.
  \begin{align}
    \Delta\bm{l}^{ref}_{s, t} &= -(\bm{l}^{comp}(\bm{f}^{data}_{t})-\bm{l}^{comp}(\bm{f}^{ref}_{t}))
  \end{align}
  As in (B), we use the assumption of (i) and calculate the software elongation of muscles.
  Since this is also a compensation term and a negative value, it is necessary to reverse the sign of the term.

  Finally, (a)-(c) in \secref{sec:proposed} represent an interesting relationship when diagrammed as shown in \figref{figure:relationship}.
  Between the original and teaching, with $\bm{l}^{ref}$ in common, there is a difference regarding $\bm{\theta}^{ref}$ and $\bm{f}^{ref}$ depending on the external force.
  Also, between the teaching and reproduction, with $\bm{\theta}^{data}$ in common, there is a difference regarding $\bm{f}^{ref}$ and $\bm{l}^{ref}$.
  In this study, we assume (i), that the muscle tension between the original and reproduction is not significantly different.
  Therefore, we can construct a structure in which one value is fixed between each operation, and the relationship between the other two values is different.
  Normally, $\bm{\theta}^{data}$ and $\bm{f}^{ref}$ are used to obtain $\bm{l}^{ref}+\Delta\bm{l}$ in reproduction (joint-based), but in this study, $\bm{l}^{ref}+\Delta\bm{l}$ is obtained under the conditions that $\bm{\theta}^{data}$ is not obtained and $\bm{f}^{data}$ is known (muscle-based).
  It is possible to understand this scheme from the relationship of \figref{figure:relationship}.

  In this study, the time stamp $t$ is an interval of 0.2 seconds.
}%
{%
  通常の軸駆動型ヒューマノイドであれば, 修正された動作の関節角度$\bm{\theta}^{data}_{t}$を取っておき, それを指令値$\bm{\theta}^{ref}_{t}$として実機に送るだけで良い.
  しかし, \secref{sec:introduction}で述べたように筋骨格ヒューマノイドは通常関節角度$\bm{\theta}^{data}_{t}$を直接測定できないため, そのような方法は現実的でない.
  ESTにより推定された関節角度$\bm{\theta}^{est}_{t}$を用いることもできるが, ESTによる推定とCTRLによる計算のニ段階を通すためMAEの誤差が蓄積してしまう(実験章において比較実験を行う).
  一方, 修正された動作の筋長$\bm{l}^{data}_{t}$を取っておき, それを指令値$\bm{l}^{ref}_{t}$として実機に送る方法を取ることが考えられる.
  しかし, \secref{subsec:basic-structure}で述べたように外力の影響は$\bm{l}^{data}_{t}$ではなくハードウェアである非線形弾性要素や筋自体の伸びに出てしまう.
  ゆえに, 測定された筋長を送っても外力を上手く反映することができない.
  そこで, 本研究では\secref{subsec:musculoskeletal-autoencoder}で述べたMusculoskeletal AutoEncoderから得られる情報を元に, 筋ベースで筋長を補正する項$\Delta\bm{l}^{ref}_{t}$を計算し, それを足しこむことで修正された動作を再生することを考える.

  元々の筋長指令$\bm{l}^{ref}_{t}$に加えて, 以下の(A)-(C)の項を加えることで動作を再生する手法を提案する.
  これは, (A) \secref{subsec:tension-limiter}の筋張力制限制御による伸び$\Delta\bm{l}^{ref}_{e, t}$, (B) 非線形弾性や筋自体のハードウェアによる伸び$\Delta\bm{l}^{ref}_{h, t}$, (C) 筋剛性制御のソフトウェアによる伸び$\Delta\bm{l}^{ref}_{s, t}$, となっている.

  (A)は非常に単純で, \equref{eq:limiter}における$\Delta\bm{l}^{ref}_{e, t}$分だけ$\bm{l}^{ref}_{t}$に加えれば良い.

  (B)はMAEを用いてハードウェアによる筋の伸びを以下のように計算する.
  \begin{align}
    \bm{\theta}^{est}_{t} &= \bm{h}_{\bm{\theta}}(\bm{f}^{data}_{t}, \bm{l}^{data}_{t})\nonumber\\
    \Delta\bm{l}^{ref}_{h, t} &= -(\bm{h}_{\bm{l}}(\bm{\theta}^{est}_{t}, \bm{f}^{data}_{t})-\bm{h}_{\bm{l}}(\bm{\theta}^{est}_{t}, \bm{f}^{ref}_{t})) \label{eq:hardware-compensation}
  \end{align}
  これはつまり, 同じ関節角度において, 筋張力を変化させたときに, どの程度ハードウェアとしての筋長が変化するかということを表している.
  この計算は, (i)重力補償に必要な筋張力と(ii)筋のハードウェア弾性の特性は, $\bm{\theta}^{ref}$と$\bm{\theta}^{data}$の間で大きく変わらないという性質を利用している.
  教示の際の外力により筋張力が高まるが, 動作を再生するときには, (i)の仮定をおいて元の$\bm{f}^{ref}$に近い筋張力を発揮するようにする.
  また, (ii)の仮定を置くことで, \equref{eq:hardware-compensation}の計算式を簡略化できている.
  実際, $\bm{h}_{\bm{l}}(\bm{\theta}, \bm{f})$は\cite{kawaharazuka2018bodyimage}のように$\bm{h}_{1}(\bm{\theta})$と$\bm{h}_{2}(\bm{\theta}, \bm{f})$に分解することができる.
  $\bm{h}_{1}$は$\bm{f}$が働いていないときの$\bm{\theta}$における筋長を表し, $\bm{h}_{2}$は$\bm{f}$が働いているときに$\bm{\theta}$を保つための筋長のハードウェアとしての伸びの補正項を表している(常に負の値を示す).
  よって, $\bm{\theta}$として同じ$\bm{\theta}^{est}_{t}$を代入することで, \equref{eq:hardware-compensation}において$\bm{h}_{1}$は打ち消し合い, その$\bm{\theta}$回りの筋のハードウェアとしての伸びを補償する項の変化$\bm{h}_{2}(\bm{\theta}^{est}_{t}, \bm{f}^{data}_{t})-\bm{h}_{2}(\bm{\theta}^{est}_{t}, \bm{f}^{ref}_{t})$, つまり筋の伸びの変化のみが計算できる.
  しかしこれはあくまで補償項であり負の値なため, 符号を逆にする必要がある.

  (C)は\equref{eq:msc}の筋剛性制御の式から以下のように筋の伸びを計算する.
  \begin{align}
    \Delta\bm{l}^{ref}_{s, t} &= -(\bm{l}^{comp}(\bm{f}^{data}_{t})-\bm{l}^{comp}(\bm{f}^{ref}_{t}))
  \end{align}
  これも(B)と同じように(i)の性質を利用して, ソフトウェアによる筋の伸びを考慮している.
  また, 補償項であり負の値なため, 符号を逆にする必要がある.

  最後に, \secref{sec:proposed}における(a)-(c)は, \figref{figure:relationship}のように図式化すると非常に面白い関係性を表している.
  originalとmodifiedの間では, $\bm{l}^{ref}$を共通として, 外力によって$\bm{\theta}^{ref}$と$\bm{f}^{ref}$の間に差が出ている.
  また, modified とreplyの間では, $\bm{\theta}^{ref}$を共通として, $\bm{f}^{ref}$と$\bm{l}^{ref}$の間に差が出ている.
  本研究では, originalとreplyの間の筋張力は大きく変わらないだろうという(i)の仮定を置くことで, originalとreplyの間に同じような関係を成り立たせている.
  ゆえに, それぞれの動作間において, 一つの値を固定して, 残りの二つの値の関係性を変化させるという構図が出来上がっている.
  通常であればreplyにおいて$\bm{\theta}^{data}$と$\bm{f}^{ref}$から$\bm{l}^{ref}+\Delta\bm{l}$を求める(Joint-based)ところだが, $\bm{\theta}^{data}$が得られず$\bm{f}^{data}$がわかっているという性質を利用することで, $\bm{l}^{ref}+\Delta\bm{l}$を求めている(Muscle-based)という図式も理解できる.

  なお, 本研究におけるタイムスタンプ$t$は0.2 sec間隔で取得する.
}%

\subsection{Comparison of Controls Reproducing the Modified Motion} \label{subsec:comparison-methods}
\switchlanguage%
{%
  The methods to be evaluated in the comparison experiment of this study are listed below.
  \begin{itemize}
    \item ALL: $\Delta\bm{l}^{ref}_{t} = \Delta\bm{l}^{ref}_{e, t}+\Delta\bm{l}^{ref}_{h, t}+\Delta\bm{l}^{ref}_{s, t}$
    \item W-HS: $\Delta\bm{l}^{ref}_{t} = \Delta\bm{l}^{ref}_{h, t}+\Delta\bm{l}^{ref}_{s, t}$
    \item W-ES: $\Delta\bm{l}^{ref}_{t} = \Delta\bm{l}^{ref}_{e, t}+\Delta\bm{l}^{ref}_{s, t}$
    \item W-HE: $\Delta\bm{l}^{ref}_{t} = \Delta\bm{l}^{ref}_{h, t}+\Delta\bm{l}^{ref}_{e, t}$
    \item W-H: $\Delta\bm{l}^{ref}_{t} = \Delta\bm{l}^{ref}_{h, t}$
    \item W-E: $\Delta\bm{l}^{ref}_{t} = \Delta\bm{l}^{ref}_{e, t}$
    \item W-S: $\Delta\bm{l}^{ref}_{t} = \Delta\bm{l}^{ref}_{s, t}$
    \item NONE: $\Delta\bm{l}^{ref}_{t} = \bm{0}$
    \item THETA: A method calculating $\bm{\theta}^{est}_{t}$ by EST and calculating $\bm{l}^{ref}_{t}$ to realize $\bm{\theta}^{est}_{t}$.
  \end{itemize}
}%
{%
  本研究の評価実験における評価する手法群を以下に列挙する.
  \begin{itemize}
    \item ALL: $\Delta\bm{l}^{ref}_{t} = \Delta\bm{l}^{ref}_{e, t}+\Delta\bm{l}^{ref}_{h, t}+\Delta\bm{l}^{ref}_{s, t}$
    \item W-HS: $\Delta\bm{l}^{ref}_{t} = \Delta\bm{l}^{ref}_{h, t}+\Delta\bm{l}^{ref}_{s, t}$
    \item W-ES: $\Delta\bm{l}^{ref}_{t} = \Delta\bm{l}^{ref}_{e, t}+\Delta\bm{l}^{ref}_{s, t}$
    \item W-HE: $\Delta\bm{l}^{ref}_{t} = \Delta\bm{l}^{ref}_{h, t}+\Delta\bm{l}^{ref}_{e, t}$
    \item W-H: $\Delta\bm{l}^{ref}_{t} = \Delta\bm{l}^{ref}_{h, t}$
    \item W-E: $\Delta\bm{l}^{ref}_{t} = \Delta\bm{l}^{ref}_{e, t}$
    \item W-S: $\Delta\bm{l}^{ref}_{t} = \Delta\bm{l}^{ref}_{s, t}$
    \item NONE: $\Delta\bm{l}^{ref}_{t} = \bm{0}$
    \item THETA: ESTにより$\bm{\theta}^{est}_{t}$を求め, CTRLによりそれを実現する$\bm{l}^{ref}_{t}$を求める手法
  \end{itemize}
}%

\section{Experiments} \label{sec:experiment}

\begin{figure}[t]
  \centering
  \includegraphics[width=0.8\columnwidth]{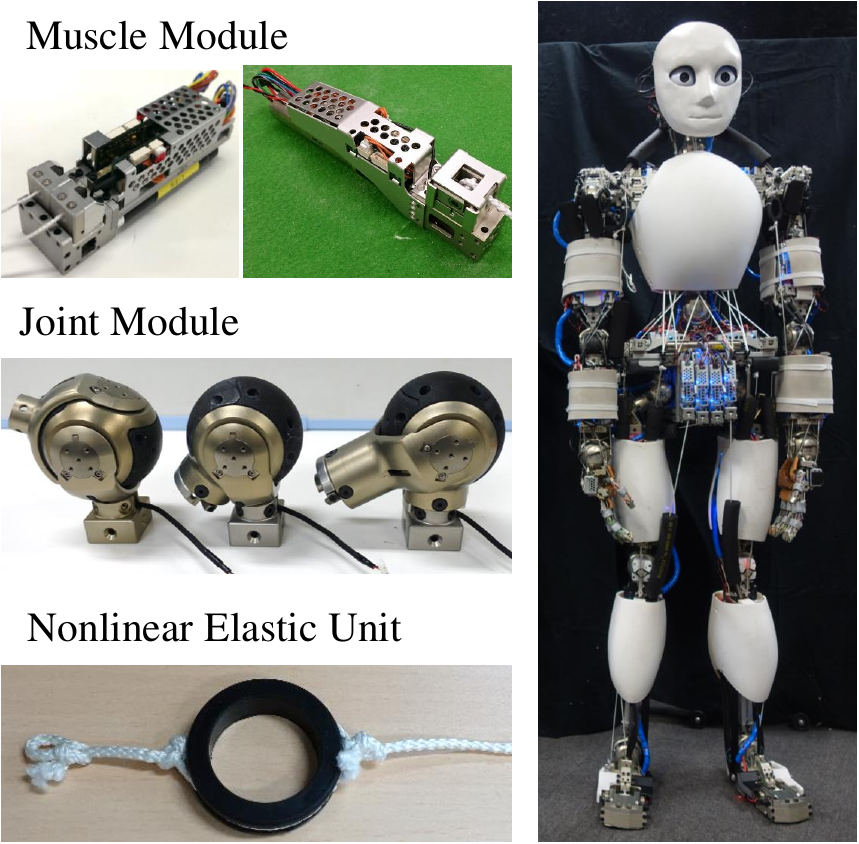}
  \caption{Experimental setup.}
  \label{figure:experimental-setup}
  %\vspace{-1.0zh}
\end{figure}

\subsection{Experimental Setup}
\switchlanguage%
{%
  In this study, we use the musculoskeletal humanoid Musashi \cite{kawaharazuka2019musashi}.
  The right figure of \figref{figure:experimental-setup} is Musashi, and the left figures show each component that constitutes it.
  A nonlinear elastic element using Grommet is arranged at the end of the muscle wire, Dyneema.
  % In this study, we mainly used three degrees of freedom (DOFs) of the shoulder and two DOFs of the elbow, which are called S-p, S-r, S-y, E-p, and E-y (S and E denote shoulder, elbow, and rpy denotes roll, pitch, and yaw).
  In this study, we mainly used three degrees of freedom (DOFs) of the shoulder and two DOFs of the elbow.
  Unlike ordinary musculoskeletal humanoids, Musashi is equipped with a mechanism to directly measure the joint angle for experimental evaluation.
  We do not use this value in our experiments, but we use it to evaluate whether or not the modified motion is accurately reproduced.
}%
{%
  本研究では筋骨格ヒューマノイドMusashi \cite{kawaharazuka2019musashi}を用いる(\figref{figure:experimental-setup}).
  Dyneemaの筋ワイヤの末端にはGrommetを用いた非線形弾性要素が配置されている.
  本研究では主に左手の肩の3自由度と肘の2自由度を用いて実験を行い, これらはS-p, S-r, S-y, E-p, E-yと呼ぶ(S, Eはshoulder, elbow, rpyはroll, pitch, yawを表す).
  また, Musashiは通常の筋骨格ヒューマノイドと違い, 実験評価のために, 関節角度を直接測定可能な機構が備わっている.
  実験の中ではこの値は用いないが, 修正された動作を正確に再現できているかどうかを評価するために用いる.
}%

\begin{figure}[t]
  \centering
  \includegraphics[width=0.8\columnwidth]{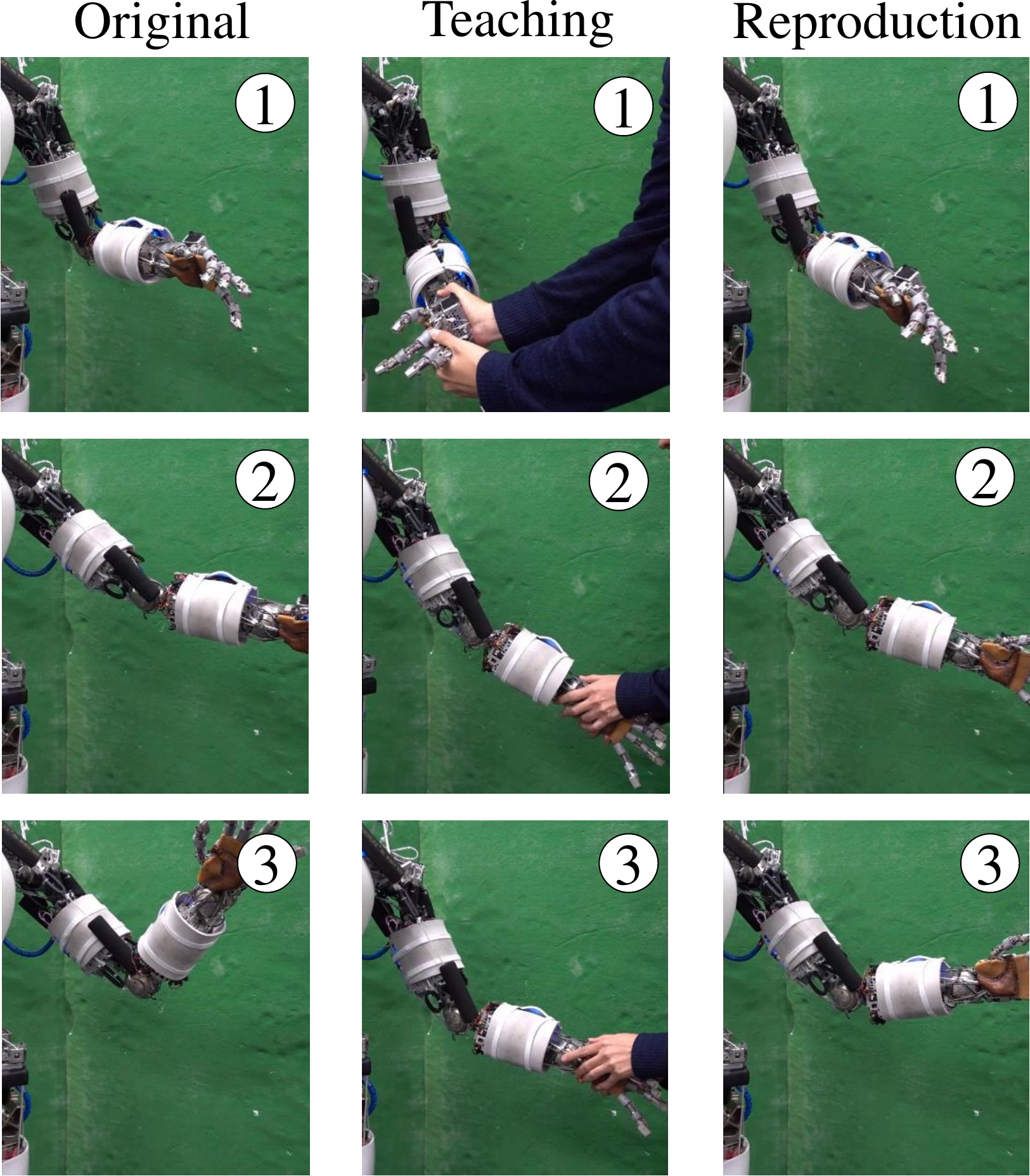}
  \caption{The procedure in the comparison experiment of motion modification methods.}
  \label{figure:comparison-experiment}
  %\vspace{-1.0zh}
\end{figure}

\begin{figure}[t]
  \centering
  \includegraphics[width=1.0\columnwidth]{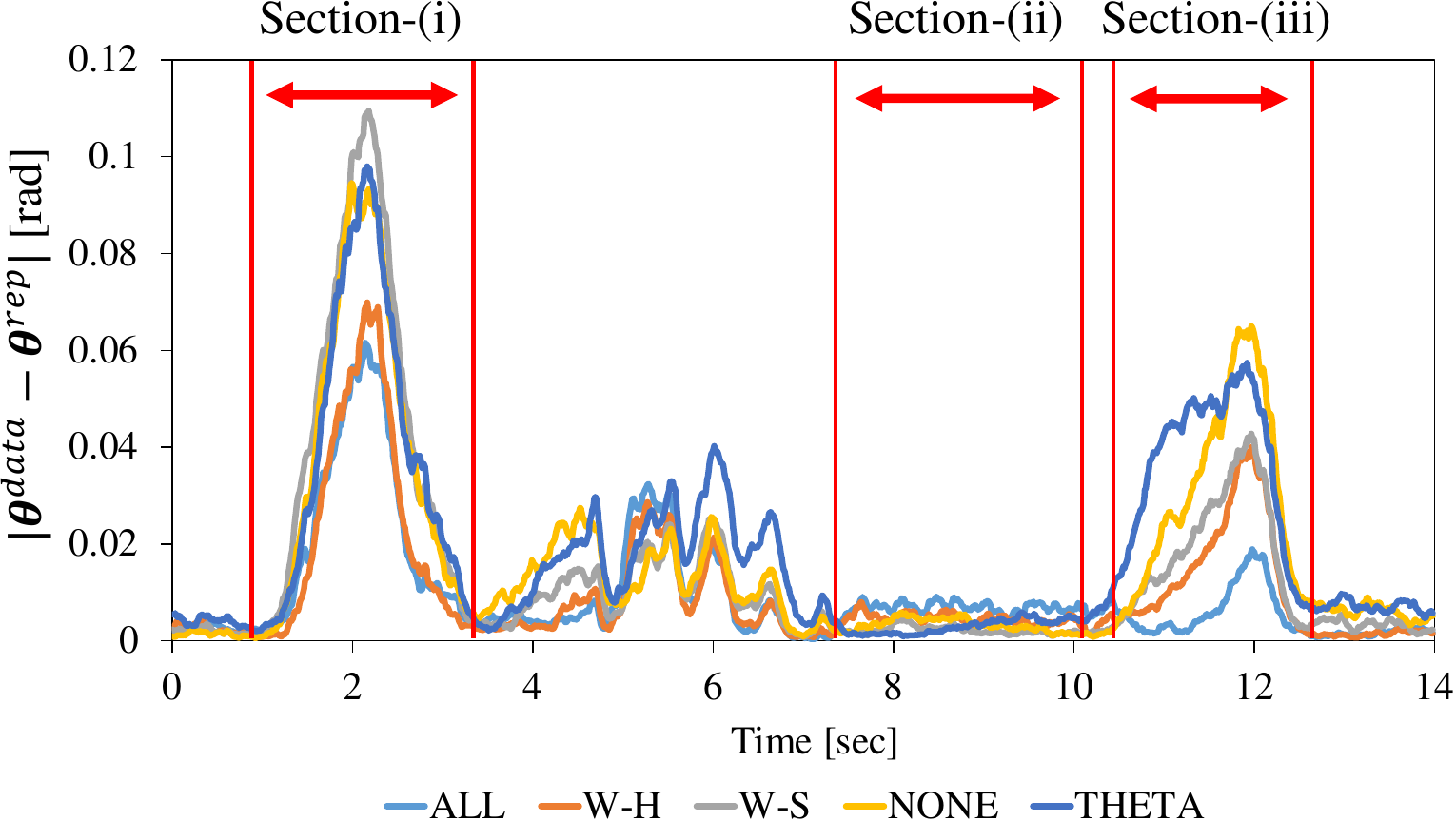}
  \caption{Transition of $|\bm{\theta}^{data}_{t}-\bm{\theta}^{rep}_{t}|$ in the comparison experiment without muscle tension limiter.}
  \label{figure:comparison-wo-result}
  %\vspace{-1.0zh}
\end{figure}

\begin{figure}[t]
  \centering
  \includegraphics[width=0.9\columnwidth]{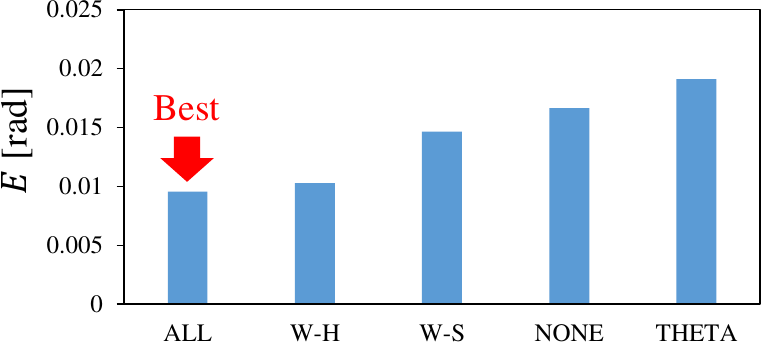}
  \caption{Comparison of evaluation value $E$ in the comparison experiment without muscle tension limiter.}
  \label{figure:comparison-wo-eval}
  %\vspace{-1.0zh}
\end{figure}

\begin{figure}[t]
  \centering
  \includegraphics[width=1.0\columnwidth]{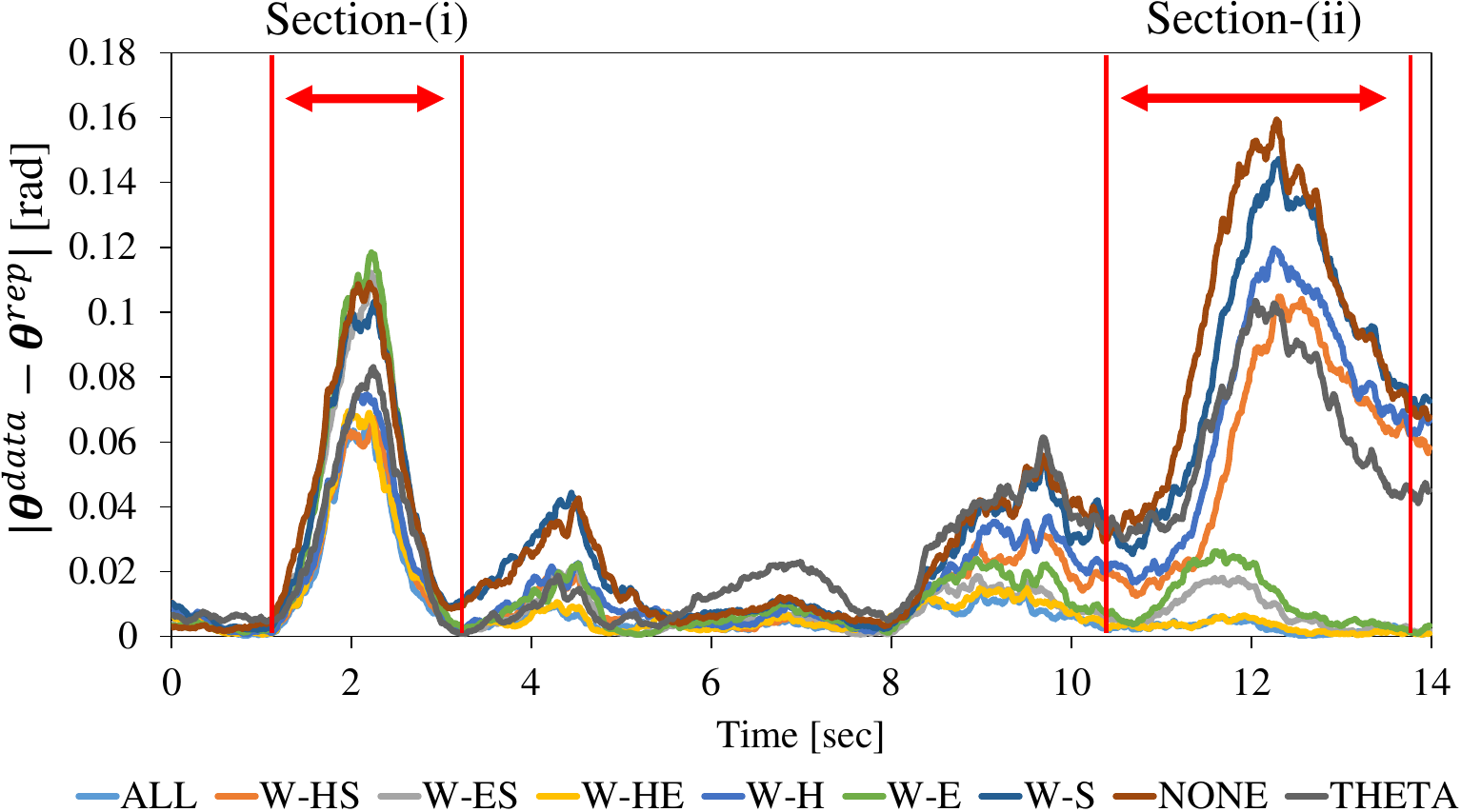}
  \caption{Transition of $|\bm{\theta}^{data}_{t}-\bm{\theta}^{rep}_{t}|$ in the comparison experiment with muscle tension limiter.}
  \label{figure:comparison-w-result}
  %\vspace{-1.0zh}
\end{figure}

\begin{figure}[t]
  \centering
  \includegraphics[width=0.9\columnwidth]{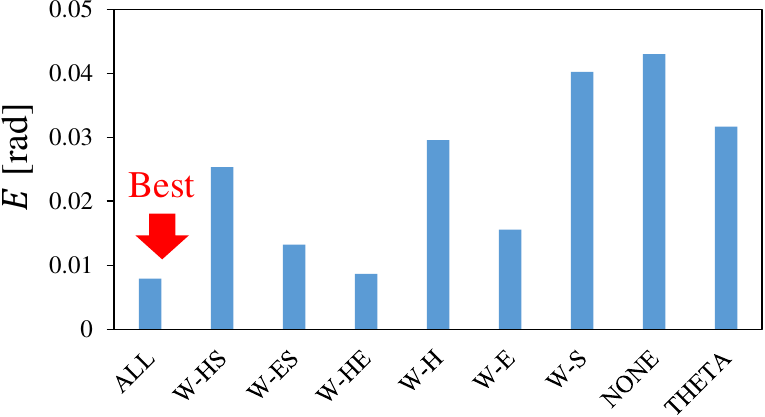}
  \caption{Comparison of evaluation value $E$ in the comparison experiment with muscle tension limiter.}
  \label{figure:comparison-w-eval}
  %\vspace{-1.0zh}
\end{figure}

\subsection{Comparison Experiment}
\switchlanguage%
{%
  In this experiment, we compare the performance of the methods presented in \secref{subsec:comparison-methods}.
  We generate the basic motion as shown in the left figures of \figref{figure:comparison-experiment}.
  Then, we modify it by applying external force as shown in the middle figures.
  Finally, the modified behavior is reproduced by the respective methods of \secref{subsec:comparison-methods} as shown in the right figures.
  Then, we compare the joint angle $\bm{\theta}^{data}_{t}$ during teaching in the middle figures and the joint angle $\bm{\theta}^{rep}_{t}$ during the reproduction in the right figures.
  $E$ expresses the average of the total time of $|\bm{\theta}^{data}_{t}-\bm{\theta}^{rep}_{t}|$.
  \figref{figure:comparison-experiment} is the experiment in the second half of this section, and reproduction shows the case of using the method ALL.
  In this experiment, the base link was moved by external force, so the comparison using the images is only a reference.

  First, the transition of $|\bm{\theta}^{data}_{t}-\bm{\theta}^{rep}_{t}|$ without muscle tension limiter of \secref{subsec:tension-limiter} is shown in \figref{figure:comparison-wo-result}.
  We focus on the intervals from Section-(i) to Section-(iii) in the 14 second operation.
  In Section-(i), the accuracy was ALL $\simeq$ W-H $>$ W-S $\simeq$ NONE $\simeq$ THETA.
  In Section-(ii), the accuracy of ALL was the worst, although they were almost the same in each case.
  In Section-(iii), the accuracy was ALL $>$ W-H $\simeq$ W-S $>$ NONE $\simeq$ THETA.
  In other words, in Section-(i), the effect of $\Delta\bm{l}^{ref}_{h, t}$ was large, while the effect of $\Delta\bm{l}^{ref}_{s, t}$ was small.
  Therefore, ALL $\simeq$ W-H and, conversely, W-S $\simeq$ NONE.
  THETA was almost the same as NONE, and its accuracy was low.
  In Section-(iii), it could be seen that the effects of $\Delta\bm{l}^{ref}_{h, t}$ and $\Delta\bm{l}^{ref}_{s, t}$ were almost the same.
  The best performance was obtained by considering the two influences of software and hardware, realized in ALL.
  On the other hand, for Section-(ii), the accuracy of ALL was the worst.
  When $|\bm{\theta}^{data}_{t}-\bm{\theta}^{rep}_{t}|$ is low overall without large external force, it is considered that errors in MAE and muscle tension measurement dominate.
  The comparison of the values of $E$ for each of these methods is shown in \figref{figure:comparison-wo-eval}.
  The accuracy results are shown as ALL $>$ W-H $>$ W-S $>$ NONE $>$ THETA, indicating that the error of ALL is about half that of THETA.

  Next, the transition of $|\bm{\theta}^{data}_{t}-\bm{\theta}^{rep}_{t}|$ in the case of using muscle tension limiter in \secref{subsec:tension-limiter} is shown in \figref{figure:comparison-w-result}.
  In this experiment, we set $f^{thre}=100$ [N].
  We similarly focus on Section-(i) and Section-(ii).
  In Section-(i), the accuracy was ALL $\simeq$ W-HS $\simeq$ W-H $\simeq$ W-H $\simeq$ THETA $>$ W-ES $\simeq$ W-E $\simeq$ W-S $\simeq$ NONE.
  In Section-(ii), the accuracy was ALL $\simeq$ W-ES $\simeq$ W-HE $\simeq$ W-E $>$ W-HS $\simeq$ W-H $\simeq$ W-S $\simeq$ NONE $\simeq$ THETA.
  In other words, in Section-(i), the effect of $\Delta\bm{l}^{ref}_{h, t}$ was the largest, whereas in Section-(ii), the effect of $\Delta\bm{l}^{ref}_{e, t}$ was the largest.
  Although THETA was good at some intervals, it was not so good when viewed as a whole.
  A comparison of the values of $E$ for each of these methods is shown in \figref{figure:comparison-w-eval}.
  The accuracy was ALL $>$ W-HE $>$ W-ES $>$ W-E $>$ W-HS $>$ THETA $>$ W-S $>$ NONE, indicating that the error of ALL was less than one third of that of THETA.
  It could also be seen that the overall degree of influence was $\Delta\bm{l}^{ref}_{e, t} >\Delta\bm{l}^{ref}_{h, t} >\Delta\bm{l}^{ref}_{s, t}$.
  The error of THETA was not much different from that of W-H.
}%
{%
  本実験では\secref{subsec:comparison-methods}で示した手法の性能を比較検討する.
  \figref{figure:comparison-experiment}の左図に示すような基本動作を記述する.
  次に, 中図のように外部から人間がそれを修正する.
  最後に, 右図のように\secref{subsec:comparison-methods}のそれぞれの手法で修正された動作を再生する.
  このとき, 中図における教示中の関節角度$\bm{\theta}^{data}_{t}$と, 下図の再生時の関節角度$\bm{\theta}^{rep}_{t}$を比較する.
  また, $|\bm{\theta}^{data}_{t}-\bm{\theta}^{rep}_{t}|$の全時間の平均を$E$とする.
 \figref{figure:comparison-experiment}は本節後半の実験であり, reproductionはALLを使った場合を示している
 本実験では外力によってbase linkが動いてしまっているため, 画像による比較は参考程度である.

  まず, \secref{subsec:tension-limiter}を用いない場合における$|\bm{\theta}^{data}_{t}-\bm{\theta}^{rep}_{t}|$の推移を\figref{figure:comparison-wo-result}に示す.
  14秒間の動作において, Section-(i)からSection-(iii)までの区間に着目する.
  Section-(i)では, 精度はALL $\simeq$ W-H $>$ W-S $\simeq$ NONE $\simeq$ THETAとなっている.
  Section-(ii)では, それぞれほとんど同じ精度であるが, ALLの精度が最も悪い.
  Section-(iii)では, 精度はALL $>$ W-H $\simeq$ W-S $>$ NONE $\simeq$ THETAとなっている.
  つまり, Section-(i)では$\Delta\bm{l}^{ref}_{h, t}$による影響が大きく, $\Delta\bm{l}^{ref}_{s, t}$による影響が小さい.
  そのため, ALL $\simeq$ W-Hとなっており, 逆にW-S $\simeq$ NONEともなっている.
  THETAはNONEとほぼ変わらず, 精度は低い.
  また, Section-(iii)では$\Delta\bm{l}^{ref}_{h, t}$と$\Delta\bm{l}^{ref}_{s, t}$の影響はほぼ同じ程度であることがわかる.
  その二つの影響を考慮したALLが最も良い性能を表している.
  これに対して, Section-(iii)では大きな差ではないにしろ, ALLの精度が最も悪い.
  あまり大きな外力が加わらず全体的に$|\bm{\theta}^{data}_{t}-\bm{\theta}^{rep}_{t}|$が低い場合は, MAEや筋張力計測の誤差の方が支配的となっていると考えられる.
  これらをまとめたそれぞれの手法の$E$の値の比較を\figref{figure:comparison-wo-eval}に示す.
  精度の結果はALL $>$ W-H $>$ W-S $>$ NONE $>$ THETAとなっており, THETAと比べるとALLの誤差は1/2程度になっていることがわかる.

  次に, \secref{subsec:tension-limiter}を用いた場合における$|\bm{\theta}^{data}_{t}-\bm{\theta}^{rep}_{t}|$の推移を\figref{figure:comparison-w-result}に示す.
  なお, $f^{thre}=100$ [N]とする.
  同様に, Section-(i), Section-(ii)の区間を設ける.
  Section-(i)では精度は, ALL $\simeq$ W-HS $\simeq$ W-HE $\simeq$ W-H $\simeq$ THETA $>$ W-ES $\simeq$ W-E $\simeq$ W-S $\simeq$ NONEとなっている.
  Section-(ii)では精度は, ALL $\simeq$ W-ES $\simeq$ W-HE $\simeq$ W-E $>$ W-HS $\simeq$ W-H $\simeq$ W-S $\simeq$ NONE $\simeq$ THETAとなっている.
  つまり, Section-(i)では$\Delta\bm{l}^{ref}_{h, t}$の影響が最も大きかったのに対して, Section-(ii)では$\Delta\bm{l}^{ref}_{e, t}$の影響が最も大きかったことが読み取れる.
  THETAは区間によっては良いこともあるが, 全体として見ると良いとは言えない.
  これらをまとめたそれぞれの手法の$E$の値の比較を\figref{figure:comparison-w-eval}に示す.
  精度の結果はALL $>$ W-HE $>$ W-ES $>$ W-E $>$ W-HS $>$ THETA $>$ W-S $>$ NONEとなっており, THETAと比べるとALLの誤差は1/3以下になっていることがわかる.
  全体的な影響度合いは$\Delta\bm{l}^{ref}_{e, t} > \Delta\bm{l}^{ref}_{h, t} > \Delta\bm{l}^{ref}_{s, t}$であることも読み取れる.
  また, THETAの誤差はW-Hと比べて大きな差は無かった.
}%

\begin{figure}[t]
  \centering
  \includegraphics[width=1.0\columnwidth]{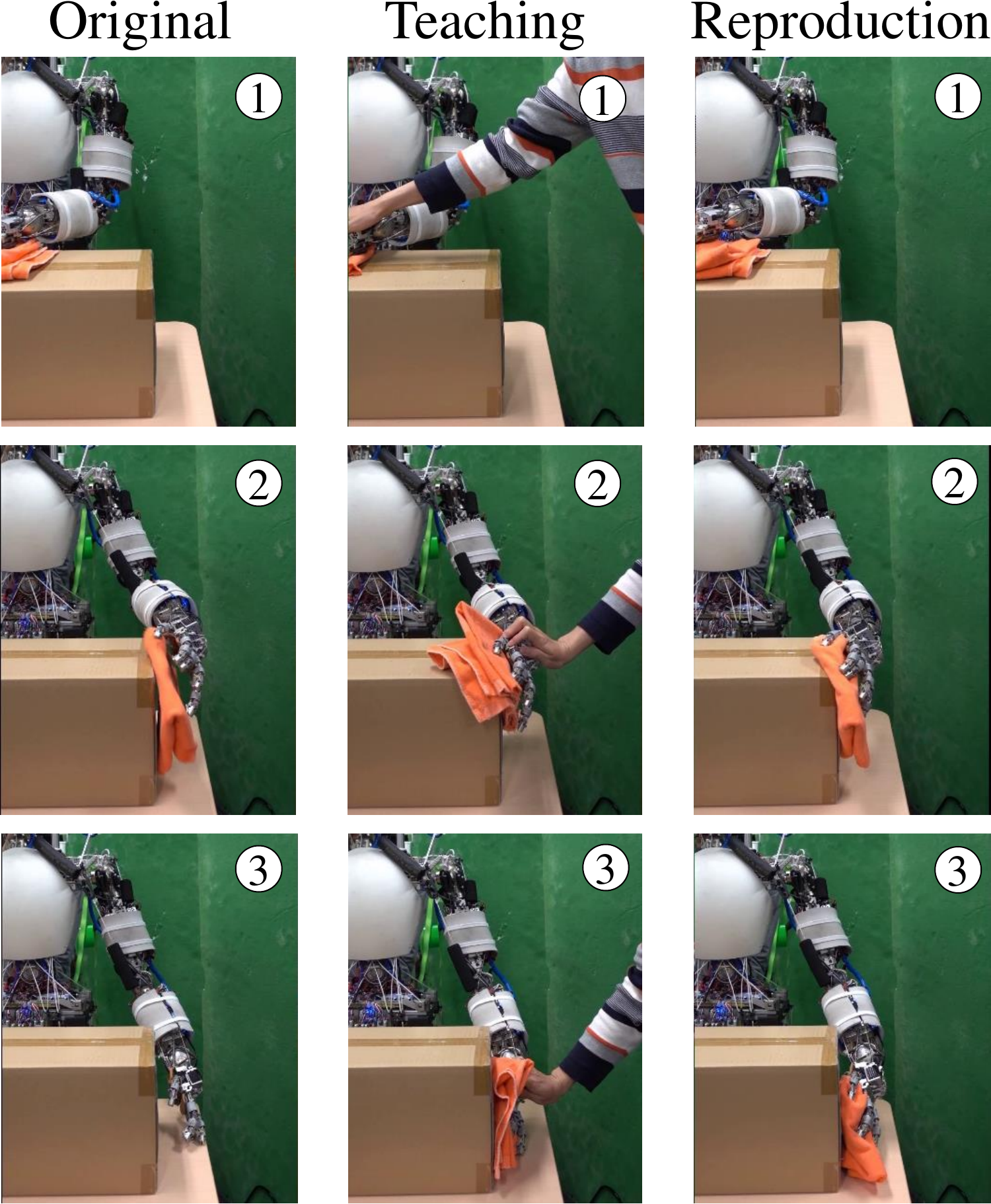}
  \caption{Box wiping experiment.}
  \label{figure:wiping-experiment}
  %\vspace{-1.0zh}
\end{figure}

\subsection{Box Wiping Experiment}
\switchlanguage%
{%
  Wiping the two planes of a box is performed using the method of this study.
  In this experiment, the muscle tension limiter is not running.
  The flow is shown in \figref{figure:wiping-experiment}.
  First, we generate the motion of wiping the box using inverse kinematics.
  However, we can see that the hand moves away from the box when wiping the vertical plane due to the error of the joint angle, and the cloth falls out of the hand.
  Next, we modify the motion so that the side of the box can be wiped correctly by applying external force during the motion.
  Finally, when the modified motion is reproduced using ALL without human guidance, we confirmed that the cloth could be wiped down to the end by pressing a hand on the side of the box.
}%
{%
  本研究の手法を用いてBoxのニ平面を拭く動作を行う.
  本実験ではmuscle tension limiterは可動させていない.
  \figref{figure:wiping-experiment}にその流れを示す.
  まず人間が逆運動学からBoxを拭く動作を生成する.
  しかし関節角度の誤差によって, 縦の平面を拭く際に手が箱から離れてしまい, 雑巾が落ちてしまっていることがわかる.
  次に, その動作中に人間がそれをガイドするように力を加え, 正しく箱の側面が拭けるように動作を修正する.
  最後に, 人間のガイド無しにALLを用いて動作を再生した結果, 箱の側面に手を押し付け, 最後まで拭くことができていることが確認できた.
}%

\begin{figure}[t]
  \centering
  \includegraphics[width=1.0\columnwidth]{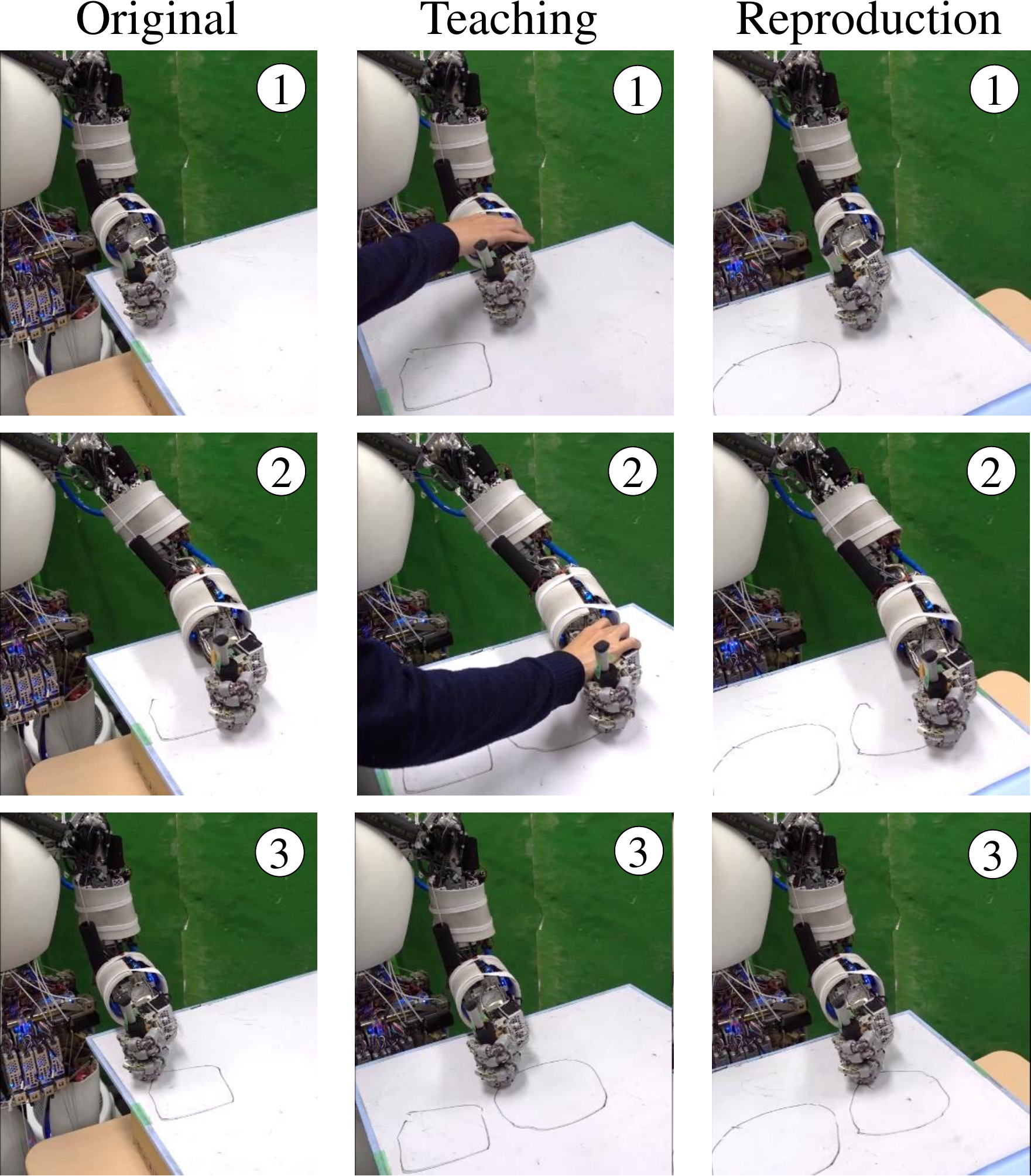}
  \caption{Drawing experiment.}
  \label{figure:drawing-experiment}
  %\vspace{-1.0zh}
\end{figure}

\begin{figure}[t]
  \centering
  \includegraphics[width=1.0\columnwidth]{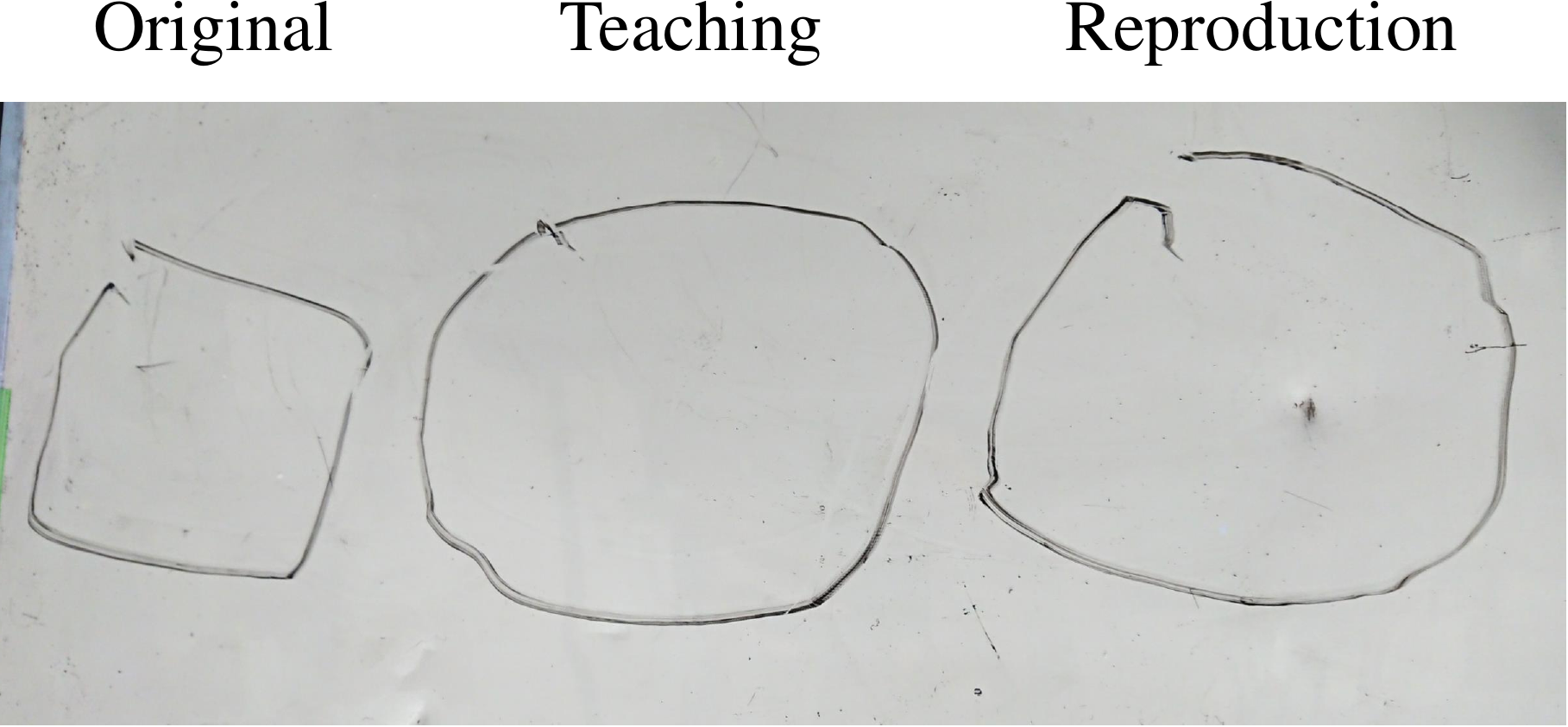}
  \caption{The result of drawn objects.}
  \label{figure:drawing-result}
  %\vspace{-1.0zh}
\end{figure}

\subsection{Drawing Experiment}
\switchlanguage%
{%
  We modify the motion of drawing a square to that of drawing a circle using our method.
  In this experiment, the muscle tension limiter is running as $f^{thre}=200$ [N].
  The flow of the experiment is shown in \figref{figure:drawing-experiment}.
  First, we generate the motion of drawing a square by using inverse kinematics.
  Next, we guide it from the outside and change the motion to the drawing of a circle.
  Finally, the modified motion is reproduced using our method, ALL.
  The final result is shown in \figref{figure:drawing-result}.
  we can see that the initial result is a rectangle, but the teaching result is a circle, and the reproduced result is a similar one.
}%
{%
  本研究の手法を用いて四角を描く動作を, 丸を描く動作に修正することを行う.
  なお, 本実験では$f^{thre}=200$ [N]としてmuscle tension limiterを可動させている.
  本実験の流れを\figref{figure:drawing-experiment}に示す.
  まず, 逆運動学を用いて四角を描く動作を生成する.
  次に, 人間がそれを外からガイドし, 丸を描く動作に変更する.
  最後に, それを本手法のALLを用いて再生する.
  最終的な結果を\figref{figure:drawing-result}に示す.
  初期はおおよそ四角を描くことができているのに対して, 教示時は丸, また, 再生時はそれと似た丸を描くことができていることがわかる.
}%

\section{Discussion} \label{sec:discussion}
\switchlanguage%
{%
  In the comparison experiment, various findings were obtained regarding the effects of the terms $\Delta\bm{l}^{ref}_{e, t}$, $\Delta\bm{l}^{ref}_{h, t}$, and $\Delta\bm{l}^{ref}_{s, t}$.
  The effect of $\Delta\bm{l}^{ref}_{e, t}$ disappears when muscle tension limiter is not used, and there are intervals in which $\Delta\bm{l}^{ref}_{h, t}$ dominates, and intervals in which the effect of $\Delta\bm{l}^{ref}_{h, t}$ and $\Delta\bm{l}^{ref}_{s, t}$ are comparable.
  This corresponds to the fact that there are intervals in which the effect of hardware is large and those in which the effect of hardware is small and is comparable to that of muscle elongation caused by software.
  Because the nonlinear elasticity of the nonlinear elastic unit degrades over time, the hardware elongation due to muscle tension may become small.
  In addition, in the case of a muscle with high muscle tension from the beginning, the hardware elongation of the muscle due to muscle tension decreases due to its nonlinearity, and it is less susceptible to the effect of external force.
  The validity of this method is demonstrated by the fact that ALL, which can take into account the effects of the hardware and software, gives the best result.
  At the same time, it is shown that the accuracy of ALL is the worst in the interval where the influence of external force is small, and that the method is able to take into account various factors and at the same time is susceptible to errors.
  Also, ALL is not able to reproduce the taught joint angle completely accurately, mainly because the assumption of MAE that the target muscle tension is completely achieved by the actual robot does not necessarily hold.
  Since MAE can only consider static factors, errors are caused by the effects of friction and hysteresis.

  In the case of using muscle tension limiter, the effects of $\Delta\bm{l}^{ref}_{e, t}$ and $\Delta\bm{l}^{ref}_{h, t}$ are found to be dominant.
  The degree of each influence varies with the intervals.
  When a strong external force is applied while the original muscle tension is high, the muscle tension is limited by $f^{thre}$ and $\Delta\bm{l}^{ref}_{e, t}$ becomes dominant.
  Conversely, if the original muscle tension is low, $\Delta\bm{l}^{ref}_{e, t}$ is not reached even when external force is applied, and $\Delta\bm{l}^{ref}_{h, t}$ becomes dominant.
  Taken as a whole, the degree of influence is $\Delta\bm{l}^{ref}_{e, t}>\Delta\bm{l}^{ref}_{h, t}>\Delta\bm{l}^{ref}_{s, t}$.
  As a whole, ALL, which can consider all elements, shows the best accuracy, demonstrating the validity of our method.
  Through the comparison experiments, we found that the accuracy of THETA is low and that it is not suitable to reproduce the modified motion of musculoskeletal humanoids directly through the estimated joint angles.

  In the box wiping and drawing experiments, the practical effectiveness of this study is shown.
  It should be noted that in the drawing experiment, we were not able to draw well when we applied this method to all the muscles.
  Due to the large friction in the elbow muscles, the error of MAE was large, the hand was lifted up when the movement was reproduced, and the pen did not touch the whiteboard.
  However, when we performed the experiment without applying this method to two muscles of the elbow, we were able to reproduce the taught movements accurately.
  In other words, the method is sensitive to the error of MAE, i.e., the estimated value of nonlinear elasticity, and we need to train it so that it can estimate the value firmly.

  Finally, the scope of this study is discussed.
  In this study, we mainly make the robot move by specifying joint angles directly or by generating simple motions with inverse kinematics and then modifying the motions.
  In the same way, we can slightly modify the motion of the robot generated by human teaching with a teaching device.
  Also, because musculoskeletal humanoids cannot measure joint angles, it is difficult to initialize the origin of muscle length.
  This method can be used to modify some of the subtle changes in the behavior when initializing it again.
  Although our method is currently designed to reproduce a single demonstration from humans, the motion may be learned from the data obtained by performing the demonstration many times in the future.
  Also, the method of incrementally modifying the behavior is expected to be developed.
}%
{%
  比較実験では, 本手法を構成する$\Delta\bm{l}^{ref}_{e, t}$, $\Delta\bm{l}^{ref}_{h, t}$, $\Delta\bm{l}^{ref}_{s, t}$の項の影響について様々な知見を得ることができた.
  muscle tension limiterを用いない場合は$\Delta\bm{l}^{ref}_{e, t}$の影響は無くなるが, $\Delta\bm{l}^{ref}_{h, t}$が支配的な区間, $\Delta\bm{l}^{ref}_{h, t}$と$\Delta\bm{l}^{ref}_{s, t}$の影響が同程度な区間が見受けられた.
  これは, 非線形弾性による影響が大きい区間と, 非線形弾性による影響が小さく, ソフトウェアによる筋の伸びと同程度である区間があることに相当する.
  非線形弾性は経年劣化するため, 徐々にクセがつき, 筋張力による伸び変化が小さくなることがあり, これが影響していると考えられる.
  また, 最初から高い筋張力がかかっている筋の場合は, その非線形性により筋張力による筋の伸びは徐々に小さくなるため, 外力による伸びの影響を受けにくくなる.
  そして, これらハードウェアとソフトウェアの影響を考慮できるALLが最も良い結果となり, 本手法の妥当性が示された.
  同時に, 外力の影響が少ない区間ではALLの精度が最も悪く, 様々な要素を考慮できると同時に, 誤差も受けやすいということが示された.
  また, 完全に正確に教示した関節角度を再生できるわけではなく, これは主に, MAEにおいて指令筋張力が実機によって完全に達成される, という過程が必ずしも成り立たないことを示している.
  MAEは静的な要素しか考慮できないため, 筋の摩擦やヒステリシスの影響で, 誤差が出てしまっている.

  muscle tension limiterを入れた場合は, $\Delta\bm{l}^{ref}_{e, t}$と$\Delta\bm{l}^{ref}_{h, t}$の影響が支配的であることがわかった.
  区間によって, それぞれの影響度合いが変化している.
  元々高い筋張力を発しているときに強い外力が加わると筋張力が$f^{thre}$によって制限され, $\Delta\bm{l}^{ref}_{e, t}$が支配的になる.
  逆に, 元々の筋張力が低い場合には外力が加わっても$\Delta\bm{l}^{ref}_{e, t}$に達さず, $\Delta\bm{l}^{ref}_{h, t}$が支配的になる.
  全体として見れば, その影響度合いは, $\Delta\bm{l}^{ref}_{e, t} > \Delta\bm{l}^{ref}_{h, t} > \Delta\bm{l}^{ref}_{s, t}$であることもわかった.
  そして全体としては, 全要素を考慮可能なALLが最も良い精度を示しており, 本手法の妥当性が示された.
  また比較実験を通して, THETAの精度は低く, 筋骨格ヒューマノイドにおいて推定関節角度を直接介して動作再生を行うのは得策ではないということがわかった.

  Box wiping実験とdrawing実験では, 本研究の実用的な有効性が示された.
  ここで特筆すべきは, drawing experiment実験では全ての筋に本手法を適用すると上手く絵を描くことはできなかったことである.
  肘の筋は大きな摩擦のためMAEの誤差が大きく, 動作再生時に手が持ち上がってしまい, ペンがホワイトボードに当たらないということが起きた.
  そこで, 肘の2本の筋のみ本手法を適用せずに実験を行うと, 正確に教示された動作を再生することができた.
  つまり, 本手法はMAEの誤差, つまり非線形弾性特性の推定値に敏感であるため, その値をしっかりと推定できるよう, 学習させる必要がある.

  最後に, 本研究の適用範囲について議論する.
  本研究では主に, 人間が関節角度を直接指定して動作をさせたり, 逆運動学で単純な動作を生成した後, それを修正するという方法を取った.
  同様に, 人間が教示デバイスを使って動かした際の動作を微妙に修正したりすることもできる.
  また, 筋骨格ヒューマノイドは関節角度を測定できないため, 筋の原点のキャリブレーションが難しい.
  そのため, キャリブレーションを行った後に動作が微妙に変化することがあるが, これらを修正することにも使用できる.
  本手法は現在人間の一度の教示を再生するものであるが, 今後, それらを何度も行ったデータから学習していくことが考えられる.
  その他, インクリメンタルに動作を修正していく方法等, 今後の発展が期待される.
}%

\section{CONCLUSION} \label{sec:conclusion}
\switchlanguage%
{%
  In this study, a method to modify the motion of the musculoskeletal humanoid by human teaching is discussed, taking advantage of the flexible body characteristics, although it is difficult to model.
  We developed a new method for reproduction of the modified motion by muscle-based compensation control, which takes into account the fact that joint angles cannot be measured and the influence of muscle tension is propagated to the hardware elasticity side of the muscle, which is different from the axis-driven type.
  Using information obtained from Musculoskeletal AutoEncoder, we confirmed through experiments that the hardware elasticity can be estimated and the modified motion can be reproduced accurately.
  Although the effects of muscle tension limiter, hardware elongation, and software elongation are different for each situation, the accuracy is the best when all of them are considered.

  In the future, we would like to investigate a method for musculoskeletal humanoids to cooperate with humans.
}%
{%
  本研究では, 筋骨格ヒューマノイドのモデル化困難ではあるが柔軟な身体の特性を活かし, 人間の教示によって動作を変化させる手法について考察した.
  軸駆動型とは異なる, 関節角度が測定できなかったり, 筋張力の影響がアクチュエータでなく筋のハードウェア弾性側に大きく伝播される特性を考慮した筋補償制御による動作再生を開発した.
  Musculoskeletal AutoEncoderから得られる情報を使い, ハードウェア弾性を推定し, 正確に修正された動作を再生することができることを実験を通して確認した.
  それぞれの状況によってmuscle tension limiterによる伸び, ハードウェアによる伸び, ソフトウェアによる伸びの影響度合いが異なるが, それらを全て統合した場合の精度が最も良くなる.

  今後, 人間と協調して作業を進める手法について深く掘り下げて実験して行きたい.
}%

{
  %\footnotesize
  %\small
  %\bibliographystyle{junsrt}
  \bibliographystyle{IEEEtran}
  \bibliography{main}
}

\end{document}